\documentclass[sigconf]{acmart}

\AtBeginDocument{%
  \providecommand\BibTeX{{%
    \normalfont B\kern-0.5em{\scshape i\kern-0.25em b}\kern-0.8em\TeX}}}

\copyrightyear{2023}
\acmYear{2023}

\acmConference[XAIFIN'23]{Workshop on Explainable AI in Finance}{November 27, 2023}{New York City, NY}
\acmPrice{15.00}
\acmISBN{978-1-4503-XXXX-X/18/06}

\setcopyright{none}
\renewcommand\footnotetextcopyrightpermission[1]{} 
\usepackage{multirow} 

\begin{document}

\title{Explainable Fraud Detection with Deep Symbolic Classification}

\author{Samantha Visbeek}
\email{SamanthaVisbeek@hotmail.com}
\affiliation{%
  \institution{RiskQuest}
  \city{Amsterdam}
  \country{the Netherlands}}

\author{Erman Acar}
\authornote{Authors contributed equally to this research.}
\email{e.acar@uva.nl}
\affiliation{%
  \institution{Universiteit van Amsterdam}
  \city{Amsterdam}
  \country{the Netherlands}
}

\author{Floris den Hengst}
\authornotemark[1]
\email{f.den.hengst@vu.nl}
\affiliation{%
  \institution{Vrije Universiteit Amsterdam}
  \city{Amsterdam}
  \country{the Netherlands}}

\renewcommand{\shortauthors}{Visbeek, Acar and den Hengst}

\begin{abstract}
There is a growing demand for explainable, transparent, and data-driven models within the domain of fraud detection.
Decisions made by the fraud detection model need to be explainable in the event of a customer dispute. Additionally, the decision-making process in the model must be transparent to win the trust of regulators, analysts, and business stakeholders. At the same time, fraud detection solutions can benefit from data due to the noisy and dynamic nature of fraud detection and the availability of large historical data sets. Finally, fraud detection is notorious for its class imbalance: there are typically several orders of magnitude more legitimate transactions than fraudulent ones.
In this paper, we present Deep Symbolic Classification (DSC), an extension of the Deep Symbolic Regression framework to classification problems. DSC casts classification as a search problem in the space of all analytic functions composed of a vocabulary of variables, constants, and operations and optimizes for an arbitrary evaluation metric directly. The search is guided by a deep neural network trained with reinforcement learning. Because the functions are mathematical expressions that are in closed-form and concise,  the model is inherently explainable both at the level of a single classification decision and at the model's decision process level. Furthermore, the class imbalance problem is successfully addressed by optimizing for metrics that are robust to class imbalance such as the F1 score. This eliminates the need for problematic oversampling and undersampling techniques that plague traditional approaches. Finally, the model allows to explicitly balance between the prediction accuracy and the explainability.
An evaluation on the PaySim data set demonstrates competitive predictive performance with state-of-the-art models, while surpassing them in terms of explainability. This establishes DSC as a promising model for fraud detection systems.
\end{abstract}





\maketitle
\section{Introduction}
\label{sec:introduction}





In recent years, the number of mobile payments has increased dramatically due to the popularity of e-Commerce, the widespread use of mobile devices, and the COVID-19 pandemic. A corresponding increase in mobile transaction fraud has been observed~\cite{bandyopadhyay2020detection}. Transaction fraud is a major challenge for the financial industry and results in monetary losses for financial institutions, account holders, and business owners. To address this issue, banks use fraud detection systems (FDS) to identify fraudulent transactions automatically and make real-time decisions on accepting or blocking a transaction. 
Deep learning has shown remarkable results in many application domains, including fraud detection~\cite{alarfaj2022credit, kim2019champion, raghavan2019fraud}. 
However, a major drawback of these models is their lack of transparency. Their black-box nature makes it difficult to justify a single decision, let alone explain their overall decision-making processes. Understanding these is necessary because (i) frauds need not only be detected, but the opportunity for fraud needs to be mitigated with, e.g., more stringent security measures, and (ii) the nature of fraud detection changes daily: new types of fraud are developed, whereas existing types fall out of favor or become impossible due to novel security measures. Furthermore, according to the European Union's 2018 General Data Protection Regulation \cite{Goodman_2017}, financial institutions are required to justify their decisions to legal authorities and customers. These requirements highlight the need for inherently transparent and explainable models. 

Explainable AI has been gaining attention in recent years, with one area of research being Symbolic Regression (SR). SR aims to find analytical (concise, closed-form) expressions that describe functional dependencies in a data set. Since an expression can be understood simply by inspection, SR can be used to create a model that is transparent and explainable. 
Recently, \citet{Petersen} proposed Deep Symbolic Regression (DSR), an approach to SR based on deep reinforcement learning. In DSR, a recurrent neural network (RNN) is trained with deep reinforcement learning on a task-specific reward function to generate expressions with high predictive power and low complexity.
DSR effectively leverages gradient-based deep learning to capture complex relationships in large data sets that are nevertheless easily described with an analytical function.



In this paper, we propose extensions to the DSR framework for the fraud detection problem. The resulting Deep Symbolic Classification (DSC) approach extends DSR with: the addition of a sigmoid layer to the output of the expressions to turn regression into binary classification, the incorporation of a threshold-based decision mechanism, and a reward function based on an accuracy metric for class-imbalanced problems.

Our approach results in a novel framework for fraud detection, characterized by the following strengths:
\begin{enumerate}
\item robust predictive performance from large-scale, high-dimensional data with the use of deep reinforcement learning,
\item analytical expressions that can be transformed into a concise set of rules and with explanations at both the decision- and the model levels,
\item intrinsically robust to highly imbalanced data without the need for problematic techniques like over- or undersampling,
\item explicitly trading off predictive power and explainability,
\item expressive power to capture non-monotonic and non-linear relationships 
without an excessive number of polynomial terms or complex neural network architectures.
\end{enumerate}

We train and evaluate DSC on the PaySim data set~\cite{paysim}. We compare our approach to de facto industry standards, including XGBoost \cite{hajek2022fraud} and show comparable predictive performance. Additionally, we evaluate the explainability of the expression obtained along two axes. Firstly, we assess whether the expression aligns with domain knowledge with an expert from the field, and find that the expression can be understood successfully. Secondly, we investigate the trade-off between explainability and predictive performance by constructing a Pareto front of performance and complexity of obtained expressions. We find that more complex expressions do not necessarily yield better predictice performance. Finally, in an investigation of the relation between expression complexity and overfitting, we find that the approach does not suffer from overfitting for simple and more complex expressions.

\section{Related Work}
\label{sec:related_work}



\noindent \textbf{Explainable Models.}
Explainable AI has gained increasing attention in recent years, particularly in fields with high societal stakes such as finance and medicine \cite{https://doi.org/10.48550/arxiv.2107.14351}. While models focusing solely on predictive performance keep surpassing the state of the art, their black-box nature prevents adoption. This is especially seen in application domains where accountability is a prerequisite and decision-making based on black-box models can have harmful consequences \cite{IR, safety}. To address this issue, the field of explainable AI has emerged \cite{xai2, alma9939747465905131, xai, exmo}. This field focuses on approaches that involve approximating a secondary model to explain the predictions made by the original black-box model. However, these approaches may be insufficiently reliable, robust, or hard to interpret themselves, which has motivated the study of inherently explainable methods \cite{alvarez-melis2018on,kumar2020problems,garreau2020explaining,rudin}. 

Furthermore, the implementation of the European Union's General Data Protection Regulation (GDPR) in 2018 has given citizens the `right to explanation' of automated decision-making models that can significantly affect them \cite{Goodman_2017}. Banks often make these decisions as part of their fraud prevention efforts. One example is the use of FDSs to block suspicious payment transactions in order to reduce losses and ensure the satisfaction of law-abiding customers.
According to \citet{inter}, bank institutions must justify their actions to customers,  anti-money laundering authorities, and legal organizations. Since fraud detection algorithms have legal, operational, strategic and ethical constraints, banks must balance explainability with predictive performance~\cite{inter}. Our approach allows for explicitly selecting a solution that is Pareto-optimal w.r.t. explainability and performance.\\


\noindent \textbf{Symbolic Regression.} 
The field of study known as Symbolic Regression (SR) aims to obtain analytical expressions that describe functional dependencies of a data set \cite{alma9940615303305131}.
Formally, given a set of characteristics $X$ and target values $\boldsymbol{y}$, with $X_{i} \in \mathbb{R}^{n}$ and $y_{i} \in \mathbb{R}$, SR aims to find a function $f: \mathbb{R}^{n} \rightarrow \mathbb{R}$ that best describes the data set, 
where $f$ is a closed-form analytical expression such that $f(X) = \boldsymbol{y} + \epsilon$. 
Essentially, the SR problem is a discrete combinatorial search for the optimal function $f^{*}$ that minimizes the distance metric $D(f(X), \boldsymbol{y})$:
\begin{equation}
    f^{*} = \underset{f}{\mathrm{argmin\,}}  D(f(X), \boldsymbol{y}).
\end{equation}
The function $f$ is an analytical expression that can be interpreted by inspection. As a result, SR is frequently used to produce models that are transparent and explainable \cite{https://doi.org/10.48550/arxiv.2107.14351}.

Conventionally, SR has emerged within the field of Genetic Programming (GP) \cite{https://doi.org/10.48550/arxiv.2107.14351}. First introduced by \citet{Koza92}, the combinatorial search problem is addressed with an algorithm inspired by the Darwinian principle of natural selection and genetic recombination. 
This process begins with the evaluation of a population of candidate solutions. Each candidate consists of a syntax tree where the leaf nodes represent features, and the internal nodes represent operators. A syntax tree represents an analytical expression and is evaluated on the training data set. In this evaluation, a predictive performance metric such as accuracy corresponds to the notion of fitness in Darwinian terminology. The fittest candidates are selected for reproduction, where their subtrees are recombined, with the goal of creating even more fit candidates.
Additionally, each node has a probability of randomly mutating, i.e. changing the node's operator. The process can be seen as a version of combinatioral discrete search for an accurate expression by training on a sufficient yet tractable number of generations.

Quite recently, \citet{sipper2022binary} showed that GP-based SR often outperforms the top machine learning methods in classification tasks, highlighting its potential for achieving high performance.
However, there are also concerns regarding the use of GP for SR, one of which is its sensitivity to hyperparameter configurations, which can result in suboptimal performance. Additionally, GP has been found to be computationally demanding and may not scale to large high-dimensional data sets with complex relationships~\cite{Petersen}.\\

\noindent \textbf{Deep Symbolic Regression.} 
To address the limitations of SR with GP, a recent work by~\cite{Petersen} has proposed a gradient-based approach to SR based on RNN and reinforcement learning, known as Deep Symbolic Regression. During training, an RNN produces analytical expressions that are then evaluated on how well they describe the data set, a measure called ``fitness''. This fitness is linked to a reward that is used to train the RNN through a risk-seeking policy gradient algorithm. The RNN adjusts the probabilities of sampling an expression according to its corresponding reward. This results in expressions that describe the data set relatively well. The authors demonstrate that DSR outperforms the included baselines on a set of benchmark problems, including Eureqa, which is considered the gold standard for symbolic regression \cite{Petersen}. 

Building on the foundation laid by DSR, \citet{mundhenk2021symbolic} have proposed a hybrid approach that combines GP and gradient-based methods. In this enhanced framework, the RNN initially generates a set of expressions (or candidate solutions). Subsequently, GP is employed as an inner optimization loop to facilitate selection, recombination, and mutation on these candidates. This enables the exploration of a more diverse solution space and enables the algorithm to escape local optima more effectively. After this GP step, the fitness of the resulting expressions is reassessed, and the RNN is then used again to generate a fresh batch of candidate solutions. This process continues iteratively, with the GP component serving as the inner optimization loop, while the neural-guided gradient-based approach operates as the outer loop. 

The hybrid approach in this study combines the strengths of both methods to enhance their respective performance. First, the integration of the RNN trained with reinforcement learning allows for improved restarts in the GP process, overcoming the limitations of random restarts that are normally associated with GP. Second, the inclusion of GP enables a more diverse exploration of the solution space, reducing the risk of being trapped in local optima. The authors demonstrate the superior performance of the hybrid DSR approach compared to vanilla DSR in various
benchmark problems \cite{mundhenk2021symbolic}.

In this study, we present a novel framework called Deep Symbolic Classification, which is a modified version of hybrid DSR, tailored specifically for the binary classification task of fraud detection. Our proposed approach incorporates a sigmoid layer into the prediction mechanism, enabling it to produce a probabilistic output suitable for classification. Additionally, we utilize the F1 score as the reward function to address the prevalent issue of high class imbalance often encountered in fraud detection scenarios. \\

\noindent \textbf{Uninterpretable Fraud Detection}
 In 2022, \citet{hajek2022fraud} proposed an XGBoost-based framework that was empirically shown to achieve state of the art (SOTA) performance on the PaySim data set. XGBoost, short for Extreme Gradient Boosting, is a decision tree ensemble method that combines multiple ML models to produce superior performance relative to that of a singular model \cite{chen2016xgboost}. It is based on the principle of sequentially adding weaker models to correct for the errors made by previous models, utilizing gradient descent to optimize the model's performance. However, one of its primary limitations is its lack of explainability, which makes it unsuitable as an FDS in practical scenarios.

In the same study, several supervised models with varying accuracy and levels of explainability were used as baselines for comparative analysis. 
These include (in decreasing order of explainability \cite{inter}) $k$-Nearest Neighbors ($k$-NN), Random Forest (RF) and Support Vector Machines (SVM).
Their findings indicated that $k$-NN and SVM are ineffective for fraud detection due to their inability to address the class imbalance. Conversely, RF performed exceptionally well on the data set, yet it is still considered a black-box model because of the high number of deep decision trees generated within it. Although some strategies have been proposed to offer the required explainability for understanding the RF model \cite{ARIA2021100094}, it is advisable to employ models that are inherently explainable instead, as previously highlighted in this section.

In this work, we conduct a comparative analysis between DSC and the aforementioned models, assessing their respective predictive performances. This enables us to illustrate how DSC measures up against models with varying degrees of explainability, as well as its performance against the state of the art on the test set. 
Note that the pre-processing of our data set differs from the procedure adopted in the work of \citet{hajek2022fraud}. Therefore, we reimplement these models according to the hyperparameters specified in the original paper to maintain consistency in our analysis. 

\section{Method}
\label{sec:methodology}



\subsection{Model implementation}
\begin{figure*}
    \centering
    \includegraphics[width=.75\textwidth]{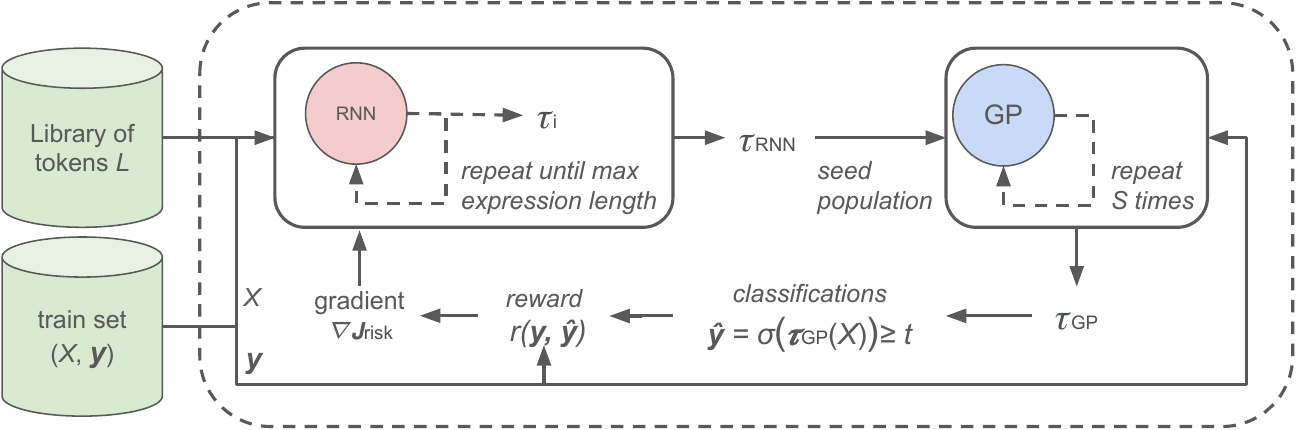}
    \caption{Train loop for deep symbolic classification. An RNN generates an expression by sampling tokens from a predefined library of tokens. The resulting expression is used to seed the population of a genetic programming process for further optimizing the expression. After optimization, classifications are generated using a sigmoid $\sigma$ and a threshold $t$. The classifications are scored with a reward function, e.g. F1 score, which is used to train the RNN in a risk-based gradient estimate.}
    \label{fig:train-loop}
\end{figure*}
We implement hybrid DSR as described in \citet{mundhenk2021symbolic}, and adjust it to make it suitable for classification problems. In this subsection, we provide a comprehensive description of our methodology for generating analytical expressions using a RNN, converting these expressions into classification models, and training the RNN using reinforcement learning techniques~\footnote{Source code available at \url{https://github.com/samanthav24/DSC_Fraud_Detection}}. Figure~\ref{fig:train-loop} details the steps described below.

\subsubsection{Generating expressions}
Each expression is represented by a binary syntax tree, following the approach as proposed by \citet{Koza92}. Each node in this tree is labeled with a token from a library of tokens. The library contains tokens that represent input features, constants, and mathematical operators. All leaf nodes of the syntax tree are labeled with tokens representing features or constants, and internal nodes represent mathematical operators. These operators can be unary (logarithm, sign, square root, etc.) or binary (summation, difference, multiplication, etc.). Within the algorithm, a syntax tree is represented as a list corresponding to a pre-order traversal of the tree, see Appendix~\ref{sec:apx:visualisation} for a visualisation. Since the arity of each mathematical operator is known upfront, a list of tokens represents a single, unique tree. The list representation brings the benefit of compatibility with existing neural network architectures that accept sequential input, including RNNs, LTSM-based models, and transformers~\cite{landajuela2022unified}.

The lists are generated from left to right, where each element is sampled from a probability distribution over all available features and mathematical operators. The probability distribution for sampling element $i$ is conditioned on the previously sampled elements $i-1, i-2, .., 0$. This conditional dependence is generated by an RNN, the outputs of which are passed through a softmax layer to obtain probabilities for each of the operators and features. 
Rather than using the current list as input, the RNN is only given the parent and sibling nodes of the previously sampled element as input. This is because the list does not capture the hierarchical structure of its syntax tree, as it was formed by preorder traversal.

In order to generate plausible expressions that make sense in the context of describing the data set, the syntax tree is subject to certain constraints: (1)\emph{the length of the sampled expression} is bound by predefined minimum and maximum values; (2) it is enforced that \emph{each pair of leaf nodes that descend from the same parent node}, should represent at least one feature: this prevents forming expressions of constants; (3) the tree has some constraints to ensure expressions that make sense mathematically: \emph{unary operators} cannot have children that are the inverse of the operator, and \emph{trigonometric operators} cannot have children that are trigonometric operators themselves. The process of generating expressions with RNN is visualized in Figure \ref{fig:sampling}.

\subsubsection{Inner optimization loop}
Entering the inner GP loop of our process, the expressions $\tau_{RNN}$ generated by the RNN serve as the initial population $\tau^{(0)}_{GP}$ for the GP algorithm. With each iteration $i$, a new population of expressions $\tau^{(i+1)}_{GP}$ is systematically constructed through processes of selection, recombination, and mutation. This iterative procedure continues until the specified number of generations, $S$, is reached. The highest-performing expressions of the final population $\tau^{(S)}_{GP}$, are selected and subsequently used for gradient update.

This hybrid methodology introduced by \citet{mundhenk2021symbolic} effectively combines the advantages of an inner GP-based optimization loop with those of an outer gradient-based optimization loop. The internal loop is similar to the standard GP with random restarts, with the distinctive addition of the RNN that offers progressively better starting populations ($\tau_{RNN} = \tau^{(0)}_{GP}$) for each successive iteration of the outer loop. In the context of the outer loop, the GP element ensures a more diverse range of expression populations. This diversity helps avoid being confined to local optima, thus facilitating a more efficient learning process.

\subsubsection{Evaluating expression}
Each expression $f$ is passed through a sigmoid function $\sigma$ to produce probabilities that are suitable for use in this binary classification problem. This allows the expression to be used to predict the likelihood that a transaction is fraudulent (1) or legitimate (0). The class prediction $\hat{y}$ of a transaction with corresponding features $\boldsymbol{x}$ is determined according to the following:
\begin{equation}
  \hat{y} =
    \begin{cases}
      1 & \text{if   } \sigma(f(\boldsymbol{x})) \geq t \\
      0 & \text{if   } \sigma(f(\boldsymbol{x})) < t
    \end{cases}       
\end{equation}
where $t$ is a given threshold. A reward is assigned to the expression, which corresponds to its performance on the data set. It is conventional in classification problems to minimize the cross-entropy loss $CE$ to increase the performance of the model.
Therefore, we define a reward function $r_{CE}$, which is the normalized inverse cross-entropy loss with respect to the ground truth classes $\boldsymbol{y}$. Specifically,
\begin{equation}
    r_{CE} = \frac{1}{1 + CE(\boldsymbol{y}, \hat{\boldsymbol{y}})} 
\end{equation}
where we add 1 to the denominator to normalize the reward. However, a major problem in the domain of fraud detection is the significant class imbalance, which often leads to low precision (i.e., too many legitimate transactions are incorrectly classified as fraudulent). To address this, a different reward function $r_{F1}$ is defined. This function is based on the F1 score, which directly optimizes the model's performance with respect to precision and recall:
\begin{equation}
    r_{F1} = \frac{2 p r}{p + r}
\end{equation}
where $p$ and $r$ are respectively the precision and recall score of the expression on the data set. We conduct a comparison between the reward functions $r_{CE}$ and $r_{F1}$ based on the performance scores of the expressions generated using these functions, for different threshold values $t \in [0.5, 0.9]$. In standard logistic regression, the threshold value is typically set to 0.5. However, due to the issue of low precision arising from the class imbalance, we increase the threshold values to improve precision.

The RNN adjusts the probability distribution for sampling elements with this reward function (a more detailed description of this process is given in Section \ref{sec:training}). 
In the context of reinforcement learning, the constituent elements of the environment, actions, episode, policy, and reward are represented by the parent and sibling nodes, the sampling of elements, the generation of an expression, the probability distribution, and the reward function, respectively.
The algorithm terminates after a given number of iterations.




\subsubsection{RNN Training}
\label{sec:training}
Since the reward function is based on the predictive power of the generated equation, rather than the parameters $\boldsymbol{\theta}$ of the RNN, it is not differentiable with respect to $\boldsymbol{\theta}$. Therefore, reinforcement learning is used to train the RNN to generate better expressions.
A naive approach to optimize the policy $p(\tau|\boldsymbol{\theta})$ (which is represented by the distribution of samples $\tau$) would be to use the standard policy gradient objective that aims to maximize the expected value of the reward. It is important to note that by maximizing an expectation, the focus is on optimizing the average performance of the generated expressions. However, in the context of fraud detection, the ultimate objective is to achieve the best possible performance of a single expression that is found during training, as it will be used as the final model. Therefore, the standard policy gradient objective may not be suitable for this purpose.

Instead, a risk seeking policy gradient objective $J_{risk}(\boldsymbol{\theta}; \varepsilon)$, is used to maximize the performance of the highest fraction of samples $\varepsilon$, at the expense of sacrificing the performance of the other generated expressions. The risk-seeking policy gradient objective \cite{Petersen} is defined as
\begin{equation}
    J_{risk}(\boldsymbol{\theta}; \varepsilon) \equiv \mathbb{E}_{\tau \sim p(\tau|\boldsymbol{\theta})} [R(\tau) | R(\tau) \geq R_{\varepsilon}(\boldsymbol{\theta})]
\end{equation}
which aims to increase the reward $R_{\varepsilon}$ of the top $\varepsilon$ fraction of samples from the distribution, while disregarding the samples that fall below this threshold. The gradient of $J_{risk}(\boldsymbol{\theta}; \varepsilon)$ is then given by
\begin{multline}
    \nabla J_{risk}(\boldsymbol{\theta}; \varepsilon) = \\ 
    \mathbb{E}_{\tau \sim p(\tau|\boldsymbol{\theta})} \big[ (R(\tau) - R_{\varepsilon}(\boldsymbol{\theta})) \cdot \nabla_{\boldsymbol{\theta}} logp(\tau|\boldsymbol{\theta}) | R(\tau) \geq R_{\varepsilon}(\boldsymbol{\theta}) \big]
\end{multline}
To compute this, we can use the standard REINFORCE Monte Carlo estimate \cite{williams1992simple} with two adjustments. First, instead of using the expected return of all samples as a baseline, we substituted it with $R_{\varepsilon}$. Second, we only include the top $\varepsilon$ fraction of samples from each batch in the gradient computation.
The resulting Monte Carlo estimate can be expressed as
\begin{multline}
    \nabla J_{risk}(\boldsymbol{\theta}; \varepsilon) \approx \\
    \frac{1}{\varepsilon N} \sum_{i = 1}^{N} \big[ R(\tau^{(i)}) - \tilde{R}_{\varepsilon}(\boldsymbol{\theta}) \big] \cdot \boldsymbol{1}_{R(\tau^{(i)}) \geq \tilde{R}_{\varepsilon}(\boldsymbol{\theta})} \nabla_{\boldsymbol{\theta}} logp(\tau^{(i)}|\boldsymbol{\theta})
\end{multline}
where $\tilde{R}_{\varepsilon}$ is the empirical $(1 - \varepsilon)$-quantile of the batch of rewards, and $\boldsymbol{1}_{s}$ takes the value 1 if the statement $s$ is true, and 0 otherwise. In this implementation, the value of $\varepsilon$ is set to 0.05, which is consistent with the approach taken in \cite{Petersen}.

\subsection{Data}
\subsubsection{The PaySim data set}
We use the popular PaySim data set~\cite{paysim}. PaySim is a data set of simulated transactions based on proprietary real transactions \cite{paysim}. Obtaining a real data set of payment transactions can be difficult due to privacy concerns. To address this issue, PaySim was developed to provide researchers with a simulated data set that exhibits statistical properties similar to a real payment transaction data set, while preserving the confidentiality of the underlying customer data. 
The data set contains \textasciitilde6.3 million transactions over a period of a month and with a fraudulent transaction rate of 0.13\%.  Columns represent various attributes associated with transactions:
\begin{itemize}
    \item step - unit of time; one step corresponds to one hour of time,
    \item type - a categorical feature with values: cash-in, cash-out, debit, payment, or transfer,
    \item amount - amount of the transaction in local currency,
    \item nameOrig - name of the customer,
    \item oldbalanceOrg - customer's balance before the transaction,
    \item newbalanceOrig - customer's balance after the transaction,
    \item nameDest - name of the recipient,
    \item oldbalanceDest - recipient's balance before the transaction,
    \item newbalanceDest - recipient's balance after the transaction,
    \item isFlaggedFraud - an indicator of whether the transaction has been flagged as fraudulent in the simulation,
    \item isFraud - an indicator of the transaction being legitimate or fraudulent,
\end{itemize}
where the column \textit{isFraud} represents the target variable, while the remaining columns are used as features in the DSC model. 

\subsubsection{Generating additional features}
Some of these features require the inclusion of pre-processing in an analytical expression targeted in this work.
The \textit{type} variable was represented with one-hot encoding. All other categorical features (\textit{nameOrig} and \textit{nameDest}) were discarded to ensure the explainability of the model.
Because information about individual customers and recipients can be essential for identifying fraud~\cite{whitrow2009transaction, bahnsen2016feature}, aggregation features were added to model the behavior of account holders. These features include the \emph{mean} and \emph{maximum} transaction amounts of the last 3 and 7 transactions of the recipient account.
Further analysis of the data set shows that only 0.15\% of the accounts participated in more than one transaction, compared to 83\% of the recipient accounts. Therefore, the aggregation of the previous 3 and 7 transaction amounts is limited to the recipient account.

The PaySim data set contains many transactions with nonzero transaction amounts and before and after balances of zero. These transactions model accounts at counterparty banks, whose balances are not known and were imputed with zero. 
 To mitigate this, the data set is enhanced with two features, \textit{externalOrig} and \textit{externalDest}, which indicate whether both balances before and after the transaction are zero, respectively, and thus are considered to belong to an external account. Furthermore, zero balances are imputed so that \textit{oldbalanceOrig} is set equal to \textit{amount} if the customer's account is external and \textit{newbalanceDest} is set equal to \textit{amount} if the recipient's account is external. This method ensures that the balances are proportional to the transaction amount. The indicator features \textit{externalOrig} and \textit{externalDest} identify such instances and differentiate between true zeros and zeros due to missing values. 
Furthermore, the inclusion of \textit{externalOrig} and \textit{externalDest} ensures that a possible relationship between fraudulent transactions and the involvement of external banks is properly considered.

The imputation of balances also mitigates a form of data leakage. A known limitation of the PaySim data set is that a model that predicts fraud if the transaction amount is equal to \textit{oldbalanceOrig} achieves exceptionally high accuracy.
This has raised concerns that the data set might have been generated according to this rule. However, the authors refute this possibility and assert that transactions recognized as fraud (as determined by the bank of the original data set from which these transactions are simulated) are likely to be canceled \cite{Lopez-Rojas_2017}, resulting in \textit{oldbalanceOrig} being set to zero. Therefore, the aforementioned imputation of \textit{oldbalanceOrig} helps circumvent this issue of data leakage.

A second form of data leakage was also mitigated in our preprocessing. According to the description of the data set, the \textit{isFlaggedFraud} feature should be True if the transaction amount exceeds 200,000. However, when analyzing the data, it becomes apparent that this condition is not met and the actual meaning of this variable remains unclear. Nevertheless, it is worth noting that almost all (99.87\%) of the transactions where \textit{isFlaggedFraud} is True, are indeed fraudulent. Due to the ambiguity surrounding the interpretation of \textit{isFlaggedFraud}, this feature is ultimately discarded.

\subsubsection{Final preprocessing steps}

Some of the baseline models we use for comparative analysis require balanced training data. To accommodate this requirement, an additional balanced training set is generated by randomly undersampling the training data. 
Details on preprocessing and all features used can be found in Appendix~\ref{sec:apx:paysim}.


\subsection{Evaluation}
We conduct a comparative analysis to assess the predictive performance of DSC in relation to the SOTA XGBoost-based method and the baseline models and hyperparameters proposed by \citet{hajek2022fraud}.
In order to assess the trade-off between expression complexity and predictive performance, we create the set of generated expressions where no other generated expression is superior in both complexity and performance \cite{Smits2005}. Such a set is typically known as the Pareto front. We define the complexity $C$ of an expression $f$ of length $T$ with sampled tokens $\tau_{i}$ as
\begin{equation}
    C(f) = \sum_{i = 0}^{T} c(\tau_{i})
\end{equation}
where $c(\tau_{i})$ is the complexity of a token $\tau_{i}$. The complexity of different types of tokens is taken from \citet{Petersen} and reproduced in Table \ref{table:comp}.

\begin{table}[tbp]
\caption{Complexity of tokens}
\label{table:comp}
\begin{tabular}{lc}
\toprule
token $\tau$                           & \multicolumn{1}{l}{complexity $c$} \\ \midrule
+, -, $\times$, feature, constant & 1                              \\
$\div$, square                                & 2                              \\
sin, cos                         & 3                              \\
exp, log, square-root                         & 4                              \\ \bottomrule
\end{tabular}
\end{table}

The optimal expression can then be determined via the \textit{elbow method}, i.e. by selecting the point at which adding more complexity to the expression does not result in a sufficient increase in the F1 score. This prevents overfitting and ensures that the expressions are not overly complex, thereby preserving explainability.

\section{Results and Discussion}
\label{sec:results}


Table \ref{table:scores} lists the performance in the test set averaged over 5 runs: the baseline classification models with and without Random Undersampling (RUS) (see Section~\ref{sec:related_work}) is compared to the best DSC configuration, i.e. with $r_{F1}$ and a threshold value of 0.8 and the best expression obtained by DSC (Equation \ref{eq:dsc_best}).
\begin{table}[tbp]
\caption{Average F1 scores on the test set. Std between parentheses if > 0.00, column-wise best in bold.}
\label{table:scores}
\begin{tabular}{lccccc}
\toprule
method                                     & accuracy      & precision     & recall        & F1 score      \\ \midrule
\multicolumn{1}{l}{RF + RUS}              & 0.93          & 0.02          & 0.93          & 0.03          \\
\multicolumn{1}{l}{XGBoost + RUS}         & 0.95          & 0.02          & \textbf{0.94} & 0.05          \\
\multicolumn{1}{l}{$k$-NN + RUS}            & 0.94          & 0.02          & 0.83          & 0.03        \\
\multicolumn{1}{l}{SVM + RUS}             & 0.95          & 0.02          & 0.70          & 0.03          \\
\multicolumn{1}{l}{RF}                    & \textbf{0.99} & \textbf{0.99} & 0.67          & 0.81          \\
\multicolumn{1}{l}{XGBoost}               & \textbf{0.99} & 0.98          & 0.70          & \textbf{0.82} \\
\multicolumn{1}{l}{DSC (average)}         & \textbf{0.99}  & 0.95 (.01)      & 0.67          & 0.78       \\
\multicolumn{1}{l}{DSC (best expression)} & \textbf{0.99} & 0.95          & 0.67          & 0.78          \\ \bottomrule
\end{tabular}
\end{table}


Table~\ref{table:scores} indicates that undersampling (RUS) does not improve results. While models trained with RUS demonstrate high recall rates, their precision values are notably low, leading to low F1 scores. This observation suggests that an excessive amount of information is lost and that models overfit to the small number of examples in the train set when using undersampling.
We note that $k$-NN and SVM demonstrate effective training only when applying RUS to the train set.
Otherwise, the training time for these models took more than 50 hours and was aborted, highlighting the complexities and resource requirements associated with highly imbalanced data sets.



In terms of F1 score, DSC demonstrates comparable performance to RF and the SOTA XGBoost model without RUS. The relatively small drop in performance compared to these baselines stems from a drop in precision and not accuracy. The precision for DSC can be considered acceptable and signifies that a relatively low number of legitimate transactions is incorrectly classified as fraudulent.
The DSC model provides inherent explainability at only a limited drop in predictive performance, making it an attractive choice for the fraud detection problem.


\subsection{Explainability}
\begin{figure}[tbp]
\centering
\includegraphics[width=0.5\textwidth]{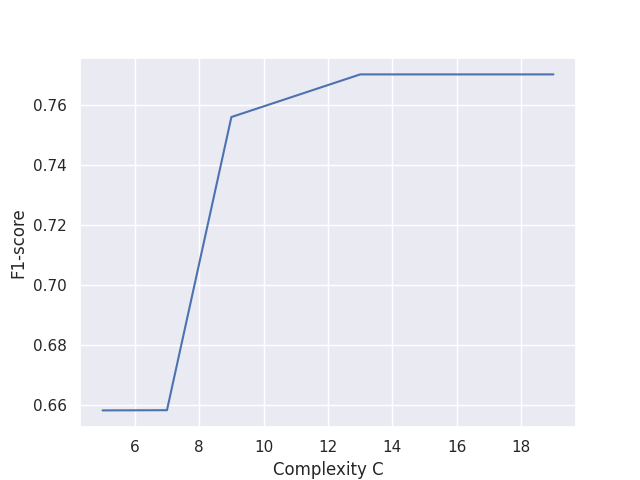}
\caption{Pareto front of predictive performance and complexity by the best DSC run ($t = 0.7$, $r_{F1}$).}
\label{fig:pareto}
\end{figure}

The best average performance was obtained with $r_{F1}$ and threshold $t = 0.8$. However, the best individual run was trained at $t=0.7$. This expression has comparable predictive performance and lower complexity. Figure~\ref{fig:pareto} shows the F1 scores of the Pareto front of the best run.
We refer to the expression $f$ with complexity level $x$, by $f_{C = x}$.
The figure indicates that the best expression, based on the F1 score, is either $f_{C=9}$ or $f_{C=13}$.
When analyzing the overfitting, one might be inclined to favor $f_{C=9}$ over $f_{C=13}$ as the performance is similar at lower complexity. However, when evaluating the expressions on the test set, it becomes apparent that $f_{C=9}$ yields a score of 0.76, while $f_{C=13}$ yields a score of $0.78$. Given that $f_{C=13}$ produces a higher F1 score on the test set, this suggests that there is no overfitting for this expression.

The key consideration now is whether the observed increase in performance justifies the corresponding increase in complexity, potentially affecting the explainability of the model. The expression that demonstrated the highest performance is given by:
\begin{multline}
\label{eq:dsc_best}
    f_{C = 13} = \sqrt{\text{externalDest} + \text{type\_cash-out}}\\
    \cdot (\text{amount} - \text{maxDest7} + \text{type\_transfer}),
\end{multline}
where \textit{maxDest7} denotes the maximum amount among the last 7 transactions (including the current amount), associated with a particular recipient.
This expression can be readily transformed into a straightforward decision rule suitable for deployment as a detection model (see Appendix~\ref{apx:rule} for details):
\textbf{classify a transaction as fraudulent, if}
\begin{itemize}
    \item type = transfer, and
    \item externalDest = True, and
    \item amount - maxDest7 > -0.15
\end{itemize}

\textbf{classify a transaction as legitimate; otherwise}\\

Compared to the best expression $f_{C = 13}$ with $f_{C = 9}$, the key difference is the absence of the square root operator and the substitution of \textit{maxDest7} with \textit{maxDest3}. However, the decision rule derived from the best expression eliminates the square-root operator, making both expressions equally explainable. The only remaining disparity lies in the utilization of either the \textit{maxDest7} or \textit{maxDest3} feature. Therefore, we consider $f_{C = 13}$ as the optimal expression.

The absence of weighted features and the lack of periodic relationships in this expression are somewhat unexpected. One plausible explanation for this finding is that the PaySim data set covers a single month, while fraudulent behavior in general exhibits seasonality over a longer period of time \cite{dal2015credit, junger2020fraud}. Hence, the complex periodicity of real-world fraud may not be present in the data set.
Furthermore, we observe that the features \textit{type\_cash-out} and \textit{type\_transfer} are present in the optimal expression. Exploratory Data Analysis confirms that all fraudulent transactions in the data set indeed fall under these two types. However, when examining the subsequent decision model derived from the expression, it becomes apparent that the model does not detect fraudulent transactions of the type "cash-out". Although the model acknowledges the significance of this type, it is likely that further training is necessary to effectively capture this specific relationship.

\subsection{Expert Interpretation}
We participated in a discussion with a senior expert in fraud detection employed at a large international bank based in the Netherlands. Our discussion focused on whether our findings align with expert knowledge and the potential applicability of our approach within the bank the expert is currently employed, considering both its performance and explainability. The expression's simplicity and ease of interpretation make it more manageable than the complex set of rules and large-scale random forests that are typically in place.
Moreover, the selected features and their relations within the expression are logically coherent.
For example, the inclusion of the feature "\textit{type}=transfer" aligns with criminal behavior. Transfers are popular for executing fraud, in contrast to other types of transactions such as payments.
Similarly, the feature "\textit{externalDest} = True" is informative. Specifically, in the event that a transaction is classified as fraudulent by an FDS, the bank may need to retrieve funds. The process of retrieving funds becomes more challenging if the transaction involves an external bank, compared to internal transfers. Fraudsters are well aware of this distinction and can exploit vulnerabilities in the system by diverting money to external institutions.

Furthermore, the requirement that the transaction amount must exceed the highest value among the previous six transactions, or differ by no more than 0.15, exemplifies an adaptation by criminals to evade detection. This adaptive behavior arises from the fact that earlier detection models successfully captured and flagged transactions that adhered to this particular behavior\footnote{It is important to acknowledge that the insights are derived from the PaySim data set do not necessarily reflect current fraudulent behavior.}. In response, fraudsters devised a new method known as "smurfing", in which multiple transactions with small amounts are used to avoid detection by the system \cite{ABN_AMRO_Bank_2019}.

Finally, in a hypothetical scenario where DSC demonstrates comparable performance on the expert's bank's internal data set, it would be regarded as a valuable addition to the FDS.
The hypothetically adequate performance of DSC and its simplicity justify its consideration for use as a component in an FDS. One could also imagine that DSC could play a role in devising mitigations for new types of fraud. 

\subsection{Impact of Hyperparameters}
Table \ref{table:r_and_t} lists the predictive performance of DSC with different reward functions $r_{CE}$ and $r_{F1}$, for various thresholds, averaged over 5 runs. It is important to note that the models were trained on the imbalanced data set and used the same threshold for both training and testing.


\begin{table}[tbp]
\caption{Mean (std) scores of DSC with different reward functions and thresholds over 5 runs, column-wise best in bold.}
\label{table:r_and_t}
\begin{tabular}{cccccc}
\toprule
reward   & $t$                & accuracy & precision & recall   & F1 score \\ \midrule
$r_{CE}$ & \multicolumn{1}{c}{0.5} & \textbf{0.99} (.00) & 0.98 (.01)  & 0.52 (.05) & 0.68 (.04) \\
         & \multicolumn{1}{c}{0.6} & \textbf{0.99} (.00) & \textbf{0.98} (.00)  & 0.50 (.00)           & 0.66 (.00) \\
         & \multicolumn{1}{c}{0.7} & \textbf{0.99} (.00) & 0.98 (.01)         & 0.53 (.07)           & 0.69 (.05) \\
         & \multicolumn{1}{c}{0.8} & \textbf{0.99} (.00) & 0.98 (.01)         & 0.55 (.07)           & 0.70 (.05) \\
         & \multicolumn{1}{c}{0.9} & \textbf{0.99} (.00) & 0.95 (.03)         & 0.59 (.08)           & 0.72 (.05) \\ \midrule
$r_{F1}$ & \multicolumn{1}{c}{0.5} & \textbf{0.99} (.00) & \textbf{0.98} (.00)  & 0.50 (.00)           & 0.66 (.00) \\
         & \multicolumn{1}{c}{0.6} & \textbf{0.99} (.00) & \textbf{0.98} (.00)  & 0.50 (.00)           & 0.66 (.00) \\
         & \multicolumn{1}{c}{0.7} & \textbf{0.99} (.00) & 0.97 (.01)         & 0.56 (.05)           & 0.71 (.03) \\
         & \multicolumn{1}{c}{0.8} & \textbf{0.99} (.00) & 0.95 (.01)         & \textbf{0.67} (.00)    & \textbf{0.78} (.00) \\
         & \multicolumn{1}{c}{0.9} & \textbf{0.99} (.00) & 0.94 (.03)         & 0.66 (.01)           & 0.78 (.01) \\ \bottomrule
\end{tabular}
\end{table}

When using $r_{CE}$ as the reward function, there appears to be a slight increase in the F1 score for higher thresholds. This increase is mostly explained by higher recall. However, in general, the recall score is relatively low: only a limited number of fraudulent transactions are detected.
In contrast, the positive relationship between the threshold and the F1 score becomes more pronounced for $r_{F1}$. Although precision slightly decreases, reward increases significantly, leading to an increase in the F1 score. This trend continues until a threshold of $t = 0.8$. Using a threshold of 0.8 yields a recall rate of 0.67, i.e. two thirds of the fraudulent transactions are detected. 

The difference in performance when using $r_{CE}$ or $r_{F1}$ can be explained as follows: the fraudulent minority class carries less weight in the calculation of the normalized inverse cross-entropy loss, resulting in minimal improvements. On the other hand, by directly optimizing the F1 score, the model ensures that the minority class is not neglected, as both precision and recall have equal importance in the calculation of the reward.

The training and testing phases use the same threshold and one might therefore assume that the threshold should not have a significant impact as feature weights can be adjusted accordingly. However, the optimization of constants includes an inner optimization loop. This loop forms a computational bottleneck. This is mitigated by faster convergence in the case of higher decision thresholds. We believe that longer run times may have resulted in comparable scores for lower decision thresholds.
A higher threshold thus serves as a practical approach to reducing computational resources without sacrificing predictive performance or model simplicity.

Furthermore, we note that recall increases with higher thresholds, while the precision remains stable or even decreases for $r_{F1}$. A high decision threshold is a common strategy to favor precision when traditional machine learning models are used on imbalanced data. 
However, in this particular case, the class imbalance is substantial, with only 0.13\% of the transactions being fraudulent. As a result, the model may exhibit overconfidence in the legitimate class, causing the sigmoid function to output probabilities that are lower than they should be. High decision thresholds force the model to predict a larger proportion of fraudulent transactions and increase the recall rate. \\

\subsection{Limitations}
Despite the promise shown by our approach, several limitations require further discussion. First, our data pre-processing and prevention of data leakage introduced noise into the aggregated features. This approach may not accurately reflect genuine behavioral patterns, and thus a larger data set could potentially improve performance. However, the computational expense of DSC raises practical considerations.
Second, the representativeness of our data is a concern. The PaySim data set, simulated from real mobile money transaction may not be representative of transaction patterns globally. This, however, is a common issue in fraud detection research as the availability of realistic data is limited due to considerations on privacy, competitivenes and systemic risk. Additionally, while our model should adapt well to evolving tactics of fraudsters we did not evaluate our approach for this due to data set limitations.
Third, in our pre-processing of balance data, we favoured mitigating leakage and the integrity of the transaction amounts at the cost of accuracy with respect to balance amount.
Fourth, due to the substantial runtime of experiments, we did not perform full cross-validation. We did perform multiple runs with varying random seeds to ensure robustness of results, however. Future studies should consider cross-validation to potentially enhance the robustness of the results even further.
Finally, DSC demonstrated a higher variance in performance relative to benchmark models such as RF and XGBoost. As our proposed framework incorporates probabilistic components, some runs may escape local optima more quickly than others.
This suggests that our approach can benefit from existing approaches to escaping local optima that have already been adopted by established techniques, including those included in the benchmark. Due to the relatively high computational expense of experimentation, we leave these improvements as future work.

\section{Conclusion}
\label{sec:conclusion}


In this work, we introduced Deep Symbolic Classification, a novel framework for explainable fraud detection in financial transactions. Our approach involves training a deep symbolic regression algorithm to generate analytical expressions with a classification-based reward function. We incorporate a sigmoid layer and a tunable decision threshold to turn regression into classification.
By using the F1 score as the reward function instead and  by setting a decision threshold of 0.8, we have effectively mitigated the challenges associated with high class imbalance, a key issue in the fraud detection domain. By taking the class imbalance problem head on, DSC eliminates the need for problematic techniques such as oversampling or undersampling.
The models generated by DSC are transparent and allow for straightforward inspection of features. In particular, our analysis has revealed that certain key features align with expert knowledge about fraudulent transactions. We observed that transaction type, intra-bank transactions, and the amount of the transaction relative to the last six transactions of the recipient were significant factors in determining fraudulent activities.

Our framework facilitates the creation of models with varying complexity and predictive performance and the creation of a Pareto front. Analysts and other stakeholders can select the model that best aligns with their desired trade-off between explainability and predictive performance from this set of optimal solutions.
In our case study, we found an optimal solution that could be transformed into a concise decision rule based on only three Boolean variables.

Elaborating further on the aspect of predictive performance, DSC exhibits slightly lower performance compared to SOTA models on the PaySim data set. 
However, DSC achieves precision and recall scores of 0.95 and 0.67, respectively, indicating a minimal occurrence of misclassified legitimate transactions and a notable ability to detect approximately two thirds of fraudulent transactions.
It is important to note that the SOTA models lack explainability and exhibit only marginally better performance, thus positioning DSC as a promising model for fraud detection. However, additional research needs to be done on different data sets and different operators to provide a definitive conclusion regarding its practical implementation in industry. \\

Regarding future directions, several areas can be explored.
Firstly, incorporating relational operators (e.g., $\geq$, $<$, $\neq$) or aggregational operators (e.g., mean, standard deviation, percentiles) in the library of tokens can help eliminate the need for manual feature engineering.

Additionally, exploring alternatives to the sigmoid function for mapping expression values to probability spaces could be fruitful. Multilayer Discriminant Classification presents an interesting option, wherein two expressions are created, one for each class, and the argmax of their weighted values determine the classification \cite{sipper2022binary}. The weights of the features in both expressions directly optimize the likelihood of each class.

Moreover, the recurrent expression generation process of DSC, trained via reinforcement learning, lacks parallelization, resulting in relatively high computation times. A potential solution is to investigate transformer-based symbolic regression, as introduced by \citet{kamienny2022end}, to address this limitation.

\begin{acks}
We kindly thank Wim Tip for sharing his expertise on fraud detection and the anonymous reviewers for their useful suggestions to improve on this work.
\end{acks}

\bibliographystyle{ACM-Reference-Format}
\bibliography{bibliography}

\appendix
\onecolumn
\section{Visualisation of sampling expressions with DSR}
\label{sec:apx:visualisation}
\begin{figure*}[h]
    \centering
    \includegraphics[width=0.8\textwidth]{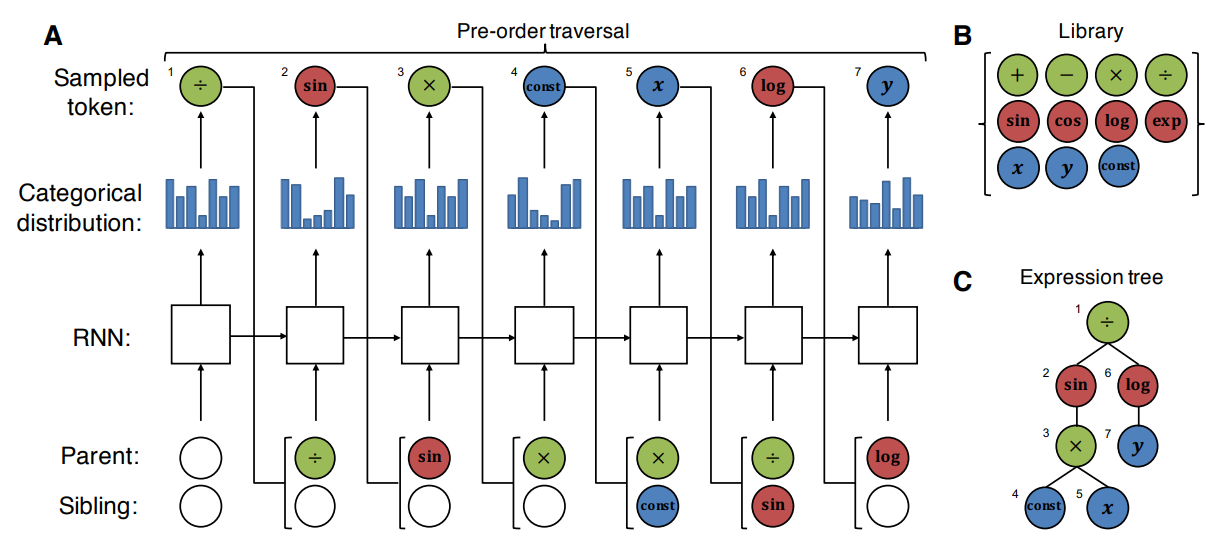}
    \caption{An example of the sampling process of (Hybrid) Deep Symbolic Regression. Image is taken from the original DSR paper by \citet{Petersen}. \textbf{(A)} Elements are sampled from a categorical distribution emitted by the RNN on the library $L$ of elements, which is given in \textbf{(B)}. 
    The parent and sibling nodes of the next element are used as the next input to the RNN. The sampling process ends when all branches reach the leaf nodes. The resulting list [$\div, \sin, \times, constant, x, \log, y$] is the preorder traversal of the syntax tree that represents the equation $\sin(cx) / \log(y)$. 
    \textbf{(C)} The syntax tree that can be reconstructed from the preorder traversal list from \textbf{(A)}.}
    \label{fig:sampling}
\end{figure*}

\section{Preprocessing the PaySim data set}
\label{sec:apx:paysim}
The following steps were taken into account to preprocess the data set:
\begin{itemize}
    \item Certain transactions in the data set exhibited non-zero amounts, but had corresponding old and new balances of zero. To address this scenario, we introduced the features \textit{externalOrig} and \textit{externalDest} for the customer and recipient accounts, respectively (please refer to Table \ref{table:feat} for further details). Following this, we performed imputation of the balances according to the following relationships:\\
    \textit{newbalanceDest} = \textit{oldbalanceDest} + \textit{amount}\\
    \textit{oldbalanceOrig} = \textit{newbalanceOrig} + \textit{amount}

    \item Additional features were obtained through aggregation techniques in the data set. Descriptions of these features are given in Table \ref{table:feat}.

    \item The features \textit{nameOrig}, \textit{nameDest} and \textit{isFlaggedFraud} were discarded.

    \item The feature \textit{type} was one-hot encoded.

    \item The data was randomly split into a training, validation, and test set which encompassed 75\%, 10\% and 15\% of the data, respectively.

    \item  A standard scaler was fitted on the numerical columns of the training set. Subsequently, the numerical columns of the training, validation, and the test set were scaled using this fitted standard scaler.

    \item  For some of the baseline models, an additional balanced training set was generated by randomly undersampling the training data. Specifically, all fraudulent transactions were retained and an equal number of legitimate transactions was randomly selected to match the count of fraudulent instances.

\end{itemize}

We here briefly describe and motivate some modeling decisions made in the experiments.
In all experiments we aim to incorporate aggregation features that encompass \emph{all} previous transactions of both the customer and the recipient, providing insight into their overall behavior patterns. 
The PaySim data set represents 30 days of transactions, which results in a major fraction of the account holders to participate in a low number of transactions. 
As a consequence, aggregation features may not accurately describe the individual's overall behavior. 
To address this issue, we assume that subsequent transactions are independent from the current transaction: they primarily reflect the individual's general behavior and exhibit similar distributions as those observed in previous (yet unseen) months.
Therefore, we include future transactions as well in certain aggregation features. Thus, for each transaction, we add characteristics that show the \emph{mean} and \emph{maximum} transaction amount over the entire data set of both the customer and recipient. This approach has a risk of data leakage, as earlier transactions may contain information from subsequent time steps through the balance features. However, we argue that future transaction information primarily reflects general user behavior and therefore does not constitute a form of data leakage. To reflect that these features model overall customer behavior and reduce the risk of data leakage even further, we add a Gaussian noise to aggregation features that contain future information.

\begin{table}[H]
\caption{Descriptions of the additional features that were added to the data set}
\label{table:feat}
\begin{tabular}{ll}
\toprule
Feature      & Description                                                                                                                                \\ \midrule
externalOrig & Boolean variable that indicates whether the customer account is likely associated with an external institution,                            \\
             & an account is considered external if both \textit{oldbalanceOrig} and \textit{oldbalanceOrig} equal zero \\
externalDest & Boolean variable that indicates whether the recipient account is likely associated with an external institution,                           \\
             & an account is considered external if both \textit{oldbalanceDest} and \textit{oldbalanceDest} equal zero \\
meanOrig     & mean value of all transaction amounts (excluding the current amount) associated with a particular customer,                                \\
             & with added Gaussian noise with $\mu = 0$ and $\sigma = 0.01*(q - m)$ where $q$ denotes the 0.75 quantile of the                            \\
             & customer's transaction amounts and $m$ represents the minimum transaction amount.                                                          \\
meanDest     & mean value of all transaction amounts (excluding the current amount) associated with a particular recipient,                               \\
             & with added Gaussian noise with $\mu = 0$ and $\sigma = 0.01*(q - m)$ where $q$ denotes the 0.75 quantile of the                            \\
             & recipient's transaction amounts and $m$ represents the minimum transaction amount.                                                         \\
maxOrig      & maximum value of all transaction amounts (excluding the current amount) associated with a particular customer,                                 \\
             & with added Gaussian noise with $\mu = 0$ and $\sigma = 0.01*(q - m)$ where $q$ denotes the 0.75 quantile of the                            \\
             & customer's transaction amounts and $m$ represents the minimum transaction amount.                                                          \\
maxDest      & maximum value of all transaction amounts (excluding the current amount) associated with a particular recipient,                                \\
             & with added Gaussian noise with $\mu = 0$ and $\sigma = 0.01*(q - m)$ where $q$ denotes the 0.75 quantile of the                            \\
             & recipient's transaction amounts and $m$ represents the minimum transaction amount.                                                         \\
meanDest3    & mean of the last 3 transaction amounts (including the current amount) associated with a particular recipient                               \\
meanDest7    & mean of the last 7 transaction amounts (including the current amount) associated with a particular recipient                               \\
maxDest3     & maximum of the last 3 transaction amounts (including the current amount) associated with a particular recipient                                \\
maxDest7     & maximum of the last 7 transaction amounts (including the current amount) associated with a particular recipient                                \\
numTransOrig & total number of transactions associated with a particular customer                                                                         \\
numTransDest & total number of transactions associated with a particular recipient                                                                        \\ \bottomrule
\end{tabular}
\end{table}

\section{Baseline model configuration}
\label{apx:baseline}
The training set was randomly undersampled to achieve a balanced training set. Both the balanced training set and the original training set were used to train the baseline models. Subsequently, these models were tested on an unbalanced test set. The parameters of the baseline models are displayed in Table \ref{table:baseline}.

\begin{table}[H]
\caption{Parameters of the baseline models}
\label{table:baseline}
\begin{tabular}{lll}
\toprule
Model   & Library      & Parameters                                                                                                \\ \midrule
$k$-NN    & scikit-learn & $k$ = 2                                                                                                \\
SVM     & scikit-learn & complexity parameter C = 1, kernel function = polynomial, gamma = 0.01                \\
RF      & scikit-learn & number of trees = 200                                                                            \\
XGBoost & XGBoost      & booster = gbtree, eta = 0.3, gamma = 0, maximum depth of a tree = 3, sampling method = uniform, \\
        &              & lambda = 1, alpha = 0        \\ \bottomrule                                                                            
\end{tabular}
\end{table}

\section{Derivation of decision rule}
\label{apx:rule}
The expression that resulted in the highest performance was:

\begin{equation}
\label{eq:best}
    f = \sqrt{\text{externalDest} + \text{type\_cash-out}} \cdot (\text{amount} - \text{maxDest7} + \text{type\_transfer}),
\end{equation}
where we have three Boolean features that either have value 0 or 1: \textit{externalDest}, \textit{type\_cash-out} and \textit{type\_transfer}. The other features \textit{amount} and \textit{maxDest7} are numerical and positive. The decision rule is defined as:
\begin{align*}
    \hat{y} = 1 (\text{fraud}), \\
    \text{if    } \sigma(f) > 0.7,
\end{align*}
as this expression was found by training DSC on a threshold $t = 0.7$. Rewriting the sigmoid $\sigma(f) = (1 + e^{-f})^{-1}$ gives us:
\begin{align*}
    \hat{y} = 1 (\text{fraud}), \\
    \text{if    } f > 0.85.
\end{align*}
So for each transaction, $f$ is calculated with the feature values of that transaction, and if $f > 0.85$, we classify that transaction as fraudulent.

It should be noted that 
\begin{equation}
\label{eq:sub}
    \text{amount} - \text{maxDest7} \leq 0,
\end{equation}
since \textit{maxDest7} is the maximum of the last 7 transaction amounts (\textbf{including} the current amount) associated with a particular recipient.

Furthermore, we know that if \textit{type\_transfer} = 0, then it must be that \textit{type\_cash-out} = 1, and vice versa, since the feature \textit{type} was one-hot encoded. There are essentially two scenarios:\\

\noindent \textbf{1. \textit{type\_transfer} = 0 and \textit{type\_cash-out} = 1.} For explainability, let us divide expression \ref{eq:best} in two parts, such that $f = A \cdot B$, where 
\begin{align*}
    A = \sqrt{\text{externalDest} + \text{type\_cash-out}}\\
    B = (\text{amount} - \text{maxDest7} + \text{type\_transfer}).
\end{align*}
Given the inequality \ref{eq:sub}, we know that $B$ must be smaller than or equal to 0. Since \textit{externalDest} is a Boolean, which makes $A$ positive, it follows that $f \leq 0$. Therefore, in the scenario where the transaction is of type \textit{ cashout}, the model will never classify the transaction as fraudulent.\\

\noindent \textbf{2. \textit{type\_transfer} = 1 and \textit{type\_cash-out} = 0.} For a transaction to be classified as fraudulent within this scenario, it is imperative that the value of \textit{externalDest} is equal to 1. Otherwise, $A$ would evaluate to 0, resulting in the overall value of $f$ being 0 due to multiplication with 0. 

Now, let us assume that \textit{ externalDest} = 1, this would reduce equation \ref{eq:best} to: $f = \text{amount} - \text{maxDest7} + 1$. Given that a transaction is considered fraudulent only if $f$ is greater than 0.85, it follows that $\text{amount} - \text{maxDest7}$ must be greater than -0.15.\\

\noindent We can now summarize our findings according to the following rules:\\

\noindent \textbf{classify a transaction as fraudulent, if}
\begin{itemize}
    \item type = transfer, and
    \item externalDest = True, and
    \item amount - maxDest\_7 > -0.15
\end{itemize}
\noindent \textbf{classify a transaction as legitimate, otherwise}

\end{document}